\pdfoutput=1
\documentclass{article}
\PassOptionsToPackage{table}{xcolor}
\usepackage{iclr2027_conference,times}

\usepackage{amsmath,amssymb,amsfonts,bm}
\usepackage{graphicx}
\usepackage{booktabs}
\usepackage{array}
\usepackage{multirow}
\usepackage{xcolor}
\usepackage{placeins}
\usepackage{algorithm}
\usepackage[noend]{algpseudocode}
\usepackage{hyperref}
\usepackage{cleveref}

\crefname{algorithm}{algorithm}{algorithms}
\Crefname{algorithm}{Algorithm}{Algorithms}
\crefname{appendix}{Appendix}{Appendices}
\Crefname{appendix}{Appendix}{Appendices}

\newcommand{\E}{\mathbb{E}}
\newcommand{\vd}{{\bm{d}}}
\newcommand{\vh}{{\bm{h}}}
\newcommand{\vu}{{\bm{u}}}
\newcommand{\vz}{{\bm{z}}}
\newcommand{\best}[1]{\textbf{#1}}
\newlength{\resultgroupwidth}

\title{Activation-Conditioned Self-Distillation}

\author{\begin{minipage}[t]{\dimexpr\textwidth-2\tabcolsep\relax}\raggedright
Zhexi Lu$^{1}$, Subhajit Chaudhury$^{2}$, Tejaswini Pedapati$^{2}$, Keerthiram Murugesan$^{2}$, \\
Lei Yu$^{1}$\\[2pt]
\mdseries $^{1}$Rensselaer Polytechnic Institute \qquad $^{2}$IBM Research
\end{minipage}}

\begin{document}

\maketitle

\begin{abstract}

On-policy self-distillation uses a model as its own teacher to provide dense supervision for reasoning, often through reference-solution conditioning. Providing privileged information does not by itself ensure
effective token-level supervision throughout long responses. We introduce Activation-Conditioned Self-Distillation (ACSD), which extracts a steering vector by contrasting activations of self-generated trajectories that
reach verified correct answers within a generation budget with those of all remaining trajectories. A frozen copy of the base model applies this vector at each prediction position, and the student learns from its next-token distributions on student-generated prefixes. Outcome
verification is used for direction construction and calibration;
distillation requires neither problem-specific reference text nor teacher parameter updates. The distilled student is used alone at inference.
On each of five models, ACSD achieves the highest mean accuracy over four mathematical benchmarks among the evaluated methods. On DeepSeek-R1-0528-Qwen3-8B, mean mathematical accuracy reaches 71.9\% and LiveCodeBench v6 pass@12 reaches 70.9\%, compared with 69.0\% and 66.3\% for the reference-conditioned OPSD baseline. Contrasts among correct trajectories also support distillation, and extracted directions
can be reused across mathematical training datasets. On fixed student trajectories, ACSD maintains more stable late-position logit-update magnitudes than OPSD.
\end{abstract}

\section{Introduction}
\label{sec:intro}

Post-training has substantially advanced the reasoning capabilities of language
models, particularly in mathematics and coding. Supervised fine-tuning (SFT)
exposes models to high-quality solution demonstrations, while reinforcement
learning with verifiable rewards (RLVR) trains on the model's own responses
using correctness signals. Group Relative Policy Optimization (GRPO; \citealp{shao2024deepseekmath}), for
instance, samples multiple responses to the same problem and assigns each an
outcome reward based on final-answer correctness. However, this reward lacks
fine-grained feedback on intermediate reasoning steps, and sampling multiple long responses significantly increases training cost.
On-policy distillation (OPD) addresses this by employing a strong teacher model to
provide dense, token-level supervision over the student's own responses: the
student first attempts a problem, and the teacher then computes a next-token
distribution at every position, conditioned on the problem and the student's
preceding tokens. The student learns by matching these distributions along its
own reasoning trajectory~\citep{agarwal2024gkd}.

\begin{figure}[t]
\centering
\includegraphics[width=\linewidth]{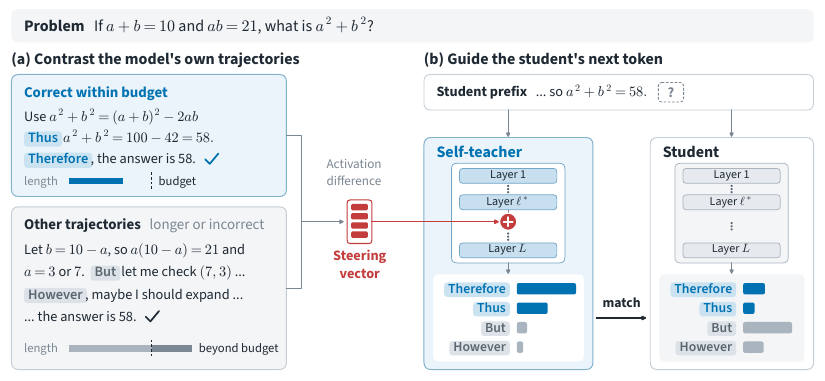}
\caption{Overview of ACSD. \textbf{(a)} We extract a steering vector from the
activation contrast between the model's own trajectories that are correct
within a generation budget and all remaining trajectories.
\textbf{(b)} The vector steers hidden states at layer $\ell^\ast$ to form the
self-teacher. Teacher and student process the same problem and
student-generated prefix, and the student learns by matching the teacher's
next-token distribution.}
\label{fig:lcsd-overview}
\end{figure}

However, OPD commonly relies on a stronger external teacher. Recent self-distillation methods instead derive a teacher from the model itself, giving it additional context that the student does not receive. OPSD conditions a frozen copy of the initial model on a correct answer or reference solution, while SDFT conditions the teacher on expert
demonstrations. Both use the model's in-context learning ability to turn this privileged context into next-token targets along the student's own
responses~\citep{zhao2026opsd,shenfeld2026sdft}. Other methods obtain conditioning information from the model's attempts and external feedback. SDPO conditions its teacher on feedback, such as execution errors, or successful attempts at the same problem~\citep{hubotter2026sdpo}.
SD-Zero trains a reviser using binary correctness feedback and then distills its next-token predictions into a
generator~\citep{he2026sdzero}. RLCSD contrasts teacher predictions conditioned on correct and incorrect solutions and uses this contrast to shape reward-based GRPO updates~\citep{pan2026rlcsd}.

Providing privileged information does not by itself ensure effective token-level supervision throughout a long response. In our fixed-trajectory analysis of OPSD, the reference-conditioned teacher's update magnitudes
decrease as the student's prefix grows
(\Cref{fig:teacher-supervision}(a)), consistent with OPSD's discussion of weaker late-token penalties~\citep{zhao2026opsd}.
This observation motivates testing whether conditioning the teacher's hidden states at every prediction position can sustain useful token-level supervision as the student's response grows longer. The model’s own attempts provide a candidate source of such conditioning: for the same problem, some reach the correct answer within a generation budget, others require longer reasoning, and still others fail. 
These attempts provide contrasts in both verified outcome and response length, including differences among successful trajectories that correctness labels alone do not capture.
Contrasting their activations offers a way to encode these differences as a reusable condition, which the teacher combines with the current student prefix when predicting each next token. We investigate whether this construction provides useful supervision throughout long responses.

We propose Activation-Conditioned Self-Distillation (ACSD), illustrated in \Cref{fig:lcsd-overview}. ACSD applies the resulting \emph{steering vector} to a frozen copy of the base model, which receives no reference answer or solution during distillation. At each prediction position, the teacher
adds this vector to its current hidden state, combining the trajectory contrast with the problem and student-generated prefix to produce a next-token distribution. The student matches these distributions on its own responses. Like OPSD, ACSD samples one response per training problem
during distillation, whereas GRPO and RLCSD use groups of responses for their reward-based updates. Its additional preparation consists of collecting and verifying trajectories, extracting the steering vector, and selecting the layer $\ell$ and strength $\alpha$ through recovery on a calibration pool. The distilled student is used alone at inference, without
activation interventions. Among the evaluated methods, ACSD achieves the highest mean accuracy across four mathematical benchmarks on each of five models and the highest LiveCodeBench v6 pass@$k$ at all evaluated
$k$ values on DeepSeek-R1-0528-Qwen3-8B (\Cref{tab:main}).

Our main contributions are as follows:
\begin{itemize}
    \item \textbf{We propose ACSD, an on-policy self-distillation method that
    constructs a teacher without reference solutions or additional teacher
    training.} We extract a steering vector from the model's own trajectory
    contrasts and apply it to a frozen teacher, which provides dense supervision
    on student-generated responses.

    \item \textbf{We analyze the role of teacher conditioning over long
    reasoning trajectories and examine the basis and reusability of the steering
    vector.} On fixed student trajectories, ACSD maintains more stable update
    magnitudes near the end of long responses than reference-conditioned OPSD.
    Contrasting shorter and longer correct trajectories also yields distillation
    gains, and the extracted directions transfer across math training datasets
    and difficulty levels.

    \item \textbf{We validate the method on mathematical reasoning and code
    generation benchmarks.} On DeepSeek-R1-0528-Qwen3-8B, ACSD achieves improvements of 4.3\% in average accuracy across four mathematical
    reasoning benchmarks and 5.2\% in LiveCodeBench v6 pass@12 over
    the base model.
\end{itemize}

\section{Preliminaries}
\label{sec:background}

\subsection{On-Policy Distillation}
\label{sec:opd-background}

Let $x$ denote a problem and $y=(y_1,\ldots,y_{|y|})$ a response sampled
from a student model $\pi_\theta$. At prediction position $t$, the prefix
$y_{<t}$ consists of the student's preceding response tokens. In on-policy
distillation, a teacher computes a next-token distribution conditioned on
$x$ and $y_{<t}$, and the student learns to match this distribution over
the vocabulary~\citep{agarwal2024gkd}. Thus, the teacher supplies a target
at each position along the student's own response, including prefixes
that differ from those the teacher would generate. Subsequent responses
are sampled from the updated student; this use of the current student's
trajectories makes the training on-policy. In self-distillation, the
teacher is derived from the model being trained. Its predictions can be
conditioned differently from the student's; for example, OPSD gives the
teacher a reference answer or solution that the student does not
receive~\citep{zhao2026opsd}.

\subsection{Activation Steering}
\label{sec:steering-background}

As a transformer processes a problem and response prefix, its layers
compute hidden-state vectors, or \emph{activations}, representing the
current context. The \emph{residual stream} is the representation passed
between transformer layers. Let $\vh_t^{(\ell)}$ denote the residual-stream
state at layer $\ell$ used to predict token $y_t$. Activation steering
modifies this state during a forward pass while keeping the model
parameters fixed~\citep{turner2023actadd,rimsky2024caa}. An additive
intervention shifts the state along a \emph{steering vector}
$\vd^{(\ell)}$; the remaining layers process the modified state to produce
the next-token distribution. The vector specifies a direction in
activation space, and its scaling determines the size of the intervention.

A common way to construct a steering vector is to contrast mean
activations from two groups of examples. For example, Contrastive
Activation Addition (CAA) uses paired prompts with contrasting
answers~\citep{rimsky2024caa}, while reasoning interventions can contrast
activations associated with annotated reasoning
behaviors~\citep{venhoff2025reasoning}. The groups, extraction positions,
and normalization rule determine the resulting direction. A fixed vector
can be reused across contexts, but its effect on the output distribution
depends on the hidden state and the subsequent network computation.
Applying steering at successive prediction positions lets the
intervention influence predictions throughout a response. The trajectory
groups, normalization, intervention scaling, and calibration used by ACSD
are specified in \Cref{sec:method}.
\section{Activation-Conditioned Self-Distillation}
\label{sec:method}

Let $\pi_0$ denote the base model, $\pi_\theta$ the student, and
$\mathcal D$ the set of distillation problems. ACSD uses a steered,
frozen copy of $\pi_0$ to supervise $\pi_\theta$ on its own trajectories.

\subsection{Motivation: From Trajectory Contrasts to a Self-Teacher}
\label{sec:reasoning-behaviors}

\begin{figure}[t]
\centering
\includegraphics[width=0.9\linewidth]{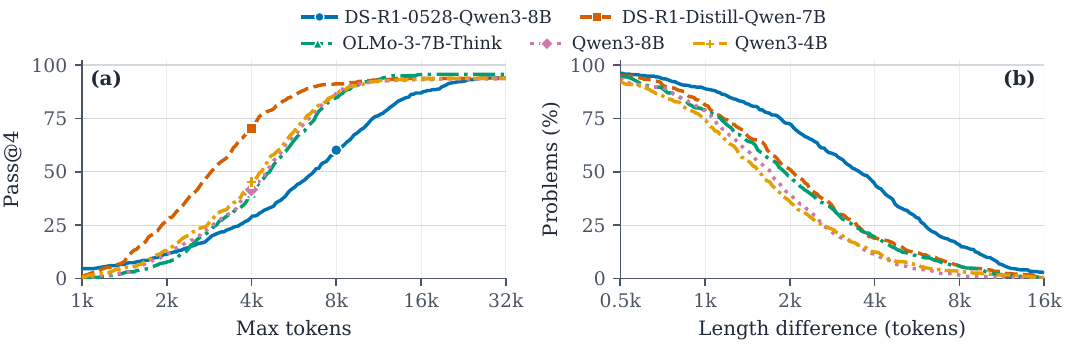}
\caption{Variation among four discovery responses per problem, with 512
problems per model. (a) Fraction of problems with at least one correct
response completed within the generation budget, computed cumulatively from
the stored responses (Pass@4). Markers indicate ACSD's budget $B$. (b)
Fraction of eligible problems whose longest
and shortest correct responses differ by at least the stated number of
tokens. Eligibility requires at least two correct responses; eligible
counts are 466, 469, 483, 476, and 476 in legend order.}
\label{fig:discovery-budget}
\end{figure}

A reasoning model can generate different trajectories for the same problem:
one attempt reaches a correct answer within a moderate number of tokens,
another succeeds only after much longer reasoning, and others fail. The
successful attempts demonstrate reasoning that the model can already
perform and provide a source of comparisons for learning from its own
behavior. \Cref{fig:discovery-budget} examines four sampled responses per
problem on 512 discovery problems for each model. Many problems have a
correct response that finishes well before the largest generation budget,
although allowing longer completions increases coverage. Even among
problems with multiple correct responses, the shortest and longest
successful trajectories can differ by thousands of tokens. Correctness
labels alone therefore do not capture the variation among successful
attempts. A generation budget provides an additional criterion: we contrast
trajectories that reach a verified correct answer within the budget with
all remaining trajectories, including longer correct attempts.

To turn this trajectory-level contrast into token-level supervision, we
need a way to incorporate it into the teacher's next-token predictions.
Activation steering provides such a mechanism: the difference between
the two groups' mean activations defines a direction that can be applied
to the teacher's hidden states. The weaker late-position updates observed
in our OPSD analysis (\Cref{fig:teacher-supervision}(a)) motivate applying
this condition at every prediction position, so that trajectory-derived
information is introduced as the student prefix grows. At each position,
the teacher processes the problem and the student's preceding tokens;
adding the direction to the resulting hidden state allows the remaining
network to combine the trajectory contrast with the current reasoning
context. The same vector can therefore influence predictions throughout
a response while producing context-dependent changes in the next-token
distribution. Whether those changes provide useful supervision is an
empirical question.

Before using this construction for distillation, we probe whether the
intervention helps the model complete prefixes from contrast-group
trajectories on problems held out from direction extraction (the
calibration pool). On DeepSeek-R1-0528-Qwen3-8B, the selected configuration
increases the fraction of verified correct continuations from 26.6\% to
66.0\% under matched sampling settings and continuation budgets
(\Cref{app:recovery}). This result shows that the extracted contrast can
improve continuation success in this setting, motivating its use to
condition a self-teacher. We construct that teacher from a frozen copy of
the base model: the student generates its own response, and the steered
teacher supplies a next-token distribution at each position along that
response (\Cref{fig:lcsd-overview}b). The student learns from these
distributions and is used alone at inference time. We evaluate the
teacher's usefulness through the distilled student's performance, since
improved recovery on the calibration pool does not by itself establish
better distillation targets. The following subsections specify the
construction, configuration selection, and student training.

\subsection{Self-Teacher Construction}
\label{sec:latent-condition}

\paragraph{Trajectory collection and labeling.}
For each discovery problem $x_i$, we sample reasoning trajectories
$r_i\sim\pi_0(\cdot\mid x_i)$. Let $V(x_i,r_i)\in\{0,1\}$ indicate
whether the final answer passes verification and $|r_i|$ the number of
generated tokens. Using a generation budget $B$, we assign the label
\begin{equation}
u_i=\mathbf 1\{V(x_i,r_i)=1\ \text{and}\ |r_i|\leq B\}.
\label{eq:rollout-label}
\end{equation}
Trajectories with $u_i=1$ form the positive group. All remaining
trajectories form the contrast group, including longer correct responses,
incorrect responses, and unfinished responses. The model-specific values
of $B$ are listed in \Cref{tab:lcsd-configs}.

\paragraph{Steering vector extraction.}
At each candidate layer $\ell$, we collect activations from the first
$\min(|r_i|,B)$ generated tokens of each trajectory. Thus, $B$ serves both
as the labeling budget and as the activation-extraction cap.
Let $\mathcal H_+^{(\ell)}$ and $\mathcal H_-^{(\ell)}$ denote the
collections of activations from the two groups. We normalize each
activation before computing the group means:
\begin{equation}
\mu_\pm^{(\ell)}
=\frac{1}{|\mathcal H_\pm^{(\ell)}|}
\sum_{\vh\in\mathcal H_\pm^{(\ell)}}
\frac{\vh}{\lVert\vh\rVert_2}.
\label{eq:activation-means}
\end{equation}
The mean pools individual token activations. Each contributes a unit
vector, while each trajectory contributes $\min(|r_i|,B)$ vectors;
normalizing activations therefore does not equalize trajectory weights.
The steering direction is the normalized difference between these means,
\begin{equation}
\vd^{(\ell)}=
\frac{\mu_+^{(\ell)}-\mu_-^{(\ell)}}
{\lVert\mu_+^{(\ell)}-\mu_-^{(\ell)}\rVert_2}.
\label{eq:direction}
\end{equation}
This unit normalization separates the direction from the intervention
strength, which we calibrate below.

\paragraph{Teacher intervention.}
A frozen copy of $\pi_0$ processes the problem and the student's preceding
tokens. At each response-prediction position $t$, we modify its residual
state at layer $\ell$ as
\begin{equation}
\widetilde{\vh}_t^{(\ell)}
=\vh_t^{(\ell)}+\alpha\lVert\vh_t^{(\ell)}\rVert_2\vd^{(\ell)},
\label{eq:steering-background}
\end{equation}
where $\alpha$ scales the intervention relative to the current
hidden-state norm. The remaining layers process the modified state to
produce the next-token distribution. Steering begins at the final prompt
state, which predicts the first response token, and continues at subsequent
response-prediction positions. The direction is fixed across positions;
its effect on predictions depends on the current problem and student
prefix.

\subsection{Configuration Selection}
\label{sec:teacher-motivation}

We select the intervention layer $\ell^\ast$ and strength $\alpha$ on the
calibration pool. For fixed prefixes from
contrast-group trajectories, we sample continuations with and without
steering under the same sampling settings and continuation budget.
\emph{Recovery} is the fraction of continuations whose final answers pass
verification; prefix construction and evaluation details are given in
\Cref{app:recovery}. After selection, we set $\vd=\vd^{(\ell^\ast)}$
and denote the resulting
teacher by $\widetilde\pi_0$. Its parameters, direction, layer, and strength
remain fixed throughout distillation. Teacher construction requires
verified self-generated trajectories and forward passes through the base
model, with no reference solution or teacher parameter updates.

\subsection{On-Policy Student Training}
\label{sec:latent-teacher}

We initialize $\pi_\theta\leftarrow\pi_0$. For each training problem
$x\sim\mathcal D$, the student generates a response
$y=(y_1,\ldots,y_{|y|})\sim\pi_\theta(\cdot\mid x)$. At position $t$,
both models process the problem $x$ and the student's preceding tokens
$y_{<t}$ to compute logits $\vz_t^S$ and $\vz_t^T$, with the
teacher applying the hidden-state intervention. At a shared distillation
temperature $T_{\mathrm d}>0$, their next-token distributions are
\begin{equation}
p_t^S=\operatorname{softmax}(\vz_t^S/T_{\mathrm d}),
\qquad
p_t^T=\operatorname{softmax}(\vz_t^T/T_{\mathrm d}).
\label{eq:teacher-student}
\end{equation}

At each position, the student learns the teacher's probability assignment
over the full vocabulary. For each entry $v$ in vocabulary $\mathcal V$, its
contribution to the forward KL divergence from teacher to student is
\begin{equation}
\ell_{t,v}=p_t^T(v)\log\frac{p_t^T(v)}{p_t^S(v)}.
\label{eq:pointwise-kl}
\end{equation}
Following the per-token pointwise clipping of \citet{zhao2026opsd}, we cap
each entry's contribution at $\tau>0$ to limit the influence of individual
vocabulary items. The training objective first clips and sums over the
vocabulary at each position, then averages over positions:
\begin{equation}
\mathcal L_{\mathrm{ACSD}}(\theta)
=\E_{x\sim\mathcal D}\E_{y\sim\pi_\theta(\cdot\mid x)}
\left[\frac{1}{|y|}\sum_{t=1}^{|y|}\sum_{v\in\mathcal V}
\min\!\left(\ell_{t,v},\tau\right)\right].
\label{eq:lcsd}
\end{equation}
Each update treats the sampled token sequence and the teacher's
distributions as constants; gradients flow only through the student
distribution $p_t^S$. Subsequent rollouts are generated by the updated
student. At inference time, only the student model is used. Hyperparameter
values ($T_{\mathrm d}$, $\tau$, and other training settings) are given in
\Cref{app:protocols}. \Cref{alg:lcsd} summarizes the procedure.

The distillation stage uses one student response per training problem, as in
OPSD. GRPO and RLCSD instead generate groups of responses for their
outcome-based updates. ACSD's additional preparation consists of rollout
collection, activation extraction, and recovery on the calibration pool to select $\ell$ and
$\alpha$; these computations are performed before student training.

\section{Experiments}
\label{sec:experiments}

\subsection{Experimental Setup}
\label{sec:setup}

\begin{table}[t]
\centering
\caption{Mathematical reasoning (Avg@12, \%) and, bottom right, LiveCodeBench
v6 code generation (pass@$k$, \%). Bold marks the best result per column
within a model; averages use unrounded scores.}
\label{tab:main}
\vspace{3pt}
\fontsize{9}{10.2}\selectfont
\setlength{\tabcolsep}{0pt}
\renewcommand{\arraystretch}{1.0}
\setlength{\resultgroupwidth}{\dimexpr(\linewidth-44pt)/3\relax}
\begin{tabular}{@{}>{\raggedright\arraybackslash}p{30pt}@{\hspace{6pt}}>{\centering\arraybackslash}p{0.195\resultgroupwidth}>{\centering\arraybackslash}p{0.195\resultgroupwidth}>{\centering\arraybackslash}p{0.24\resultgroupwidth}>{\centering\arraybackslash}p{0.195\resultgroupwidth}>{\centering\arraybackslash}p{0.175\resultgroupwidth}@{\hspace{4pt}}>{\centering\arraybackslash}p{0.195\resultgroupwidth}>{\centering\arraybackslash}p{0.195\resultgroupwidth}>{\centering\arraybackslash}p{0.24\resultgroupwidth}>{\centering\arraybackslash}p{0.195\resultgroupwidth}>{\centering\arraybackslash}p{0.175\resultgroupwidth}@{\hspace{4pt}}>{\centering\arraybackslash}p{0.195\resultgroupwidth}>{\centering\arraybackslash}p{0.195\resultgroupwidth}>{\centering\arraybackslash}p{0.24\resultgroupwidth}>{\centering\arraybackslash}p{0.195\resultgroupwidth}>{\centering\arraybackslash}p{0.175\resultgroupwidth}@{}}
\toprule
 & \multicolumn{5}{c}{\textbf{DS-R1-0528-Qwen3-8B}} & \multicolumn{5}{c}{\textbf{DS-R1-Distill-Qwen-7B}} & \multicolumn{5}{c}{\textbf{OLMo-3-7B-Think}} \\
\cmidrule(lr){2-6} \cmidrule(lr){7-11} \cmidrule(lr){12-16}
\multirow{2}{*}{Method} & {\fontsize{8}{9}\selectfont AIME} & {\fontsize{8}{9}\selectfont AIME} & {\fontsize{8}{9}\selectfont HMMT} & {\fontsize{8}{9}\selectfont AIME} & \multirow{2}{*}{Avg.} & {\fontsize{8}{9}\selectfont AIME} & {\fontsize{8}{9}\selectfont AIME} & {\fontsize{8}{9}\selectfont HMMT} & {\fontsize{8}{9}\selectfont AIME} & \multirow{2}{*}{Avg.} & {\fontsize{8}{9}\selectfont AIME} & {\fontsize{8}{9}\selectfont AIME} & {\fontsize{8}{9}\selectfont HMMT} & {\fontsize{8}{9}\selectfont AIME} & \multirow{2}{*}{Avg.} \\
 & 24 & 25 & 25 & 26 &  & 24 & 25 & 25 & 26 &  & 24 & 25 & 25 & 26 &  \\
\midrule
Base & 78.3 & 70.6 & 49.7 & 71.7 & 67.6 & 52.5 & 37.8 & 25.0 & 46.1 & 40.4 & 72.5 & 68.3 & 44.4 & 71.1 & 64.1 \\
SFT & 77.2 & 71.4 & 50.0 & 74.2 & 68.2 & 52.2 & 38.6 & 25.0 & 45.6 & 40.3 & 71.4 & 63.1 & 40.3 & 63.6 & 59.6 \\
GRPO & 79.2 & 73.9 & \best{55.8} & 74.2 & 70.8 & 56.1 & 41.4 & 26.4 & 48.3 & 43.1 & 74.7 & 68.6 & 46.9 & 71.4 & 65.4 \\
OPSD & 78.9 & 71.4 & 51.1 & 74.7 & 69.0 & \best{56.4} & 40.0 & 25.6 & \best{51.4} & 43.3 & 75.8 & 68.1 & 46.7 & 71.4 & 65.5 \\
RLCSD & 77.8 & 73.3 & 53.6 & 71.1 & 69.0 & 55.8 & 41.1 & 26.4 & 49.2 & 43.1 & 75.6 & 67.8 & 47.8 & \best{72.5} & 65.9 \\
\rowcolor{black!6}
ACSD & \best{80.8} & \best{75.0} & 55.3 & \best{76.7} & \best{71.9} & \best{56.4} & \best{41.7} & \best{27.8} & 50.3 & \best{44.0} & \best{76.1} & \best{68.9} & \best{48.1} & 71.9 & \best{66.3} \\
\bottomrule
\end{tabular}
\par\vspace{3pt}
\begin{tabular}{@{}>{\raggedright\arraybackslash}p{30pt}@{\hspace{6pt}}>{\centering\arraybackslash}p{0.195\resultgroupwidth}>{\centering\arraybackslash}p{0.195\resultgroupwidth}>{\centering\arraybackslash}p{0.24\resultgroupwidth}>{\centering\arraybackslash}p{0.195\resultgroupwidth}>{\centering\arraybackslash}p{0.175\resultgroupwidth}@{\hspace{4pt}}>{\centering\arraybackslash}p{0.195\resultgroupwidth}>{\centering\arraybackslash}p{0.195\resultgroupwidth}>{\centering\arraybackslash}p{0.24\resultgroupwidth}>{\centering\arraybackslash}p{0.195\resultgroupwidth}>{\centering\arraybackslash}p{0.175\resultgroupwidth}@{\hspace{4pt}}*{4}{>{\centering\arraybackslash}p{.25\resultgroupwidth}}@{}}
\toprule
 & \multicolumn{5}{c}{\textbf{Qwen3-8B}} & \multicolumn{5}{c}{\textbf{Qwen3-4B}} & \multicolumn{4}{c}{\textbf{DS-R1-0528-Qwen3-8B}} \\
\cmidrule(lr){2-6} \cmidrule(lr){7-11} \cmidrule(lr){12-15}
\multirow{2}{*}{Method} & {\fontsize{8}{9}\selectfont AIME} & {\fontsize{8}{9}\selectfont AIME} & {\fontsize{8}{9}\selectfont HMMT} & {\fontsize{8}{9}\selectfont AIME} & \multirow{2}{*}{Avg.} & {\fontsize{8}{9}\selectfont AIME} & {\fontsize{8}{9}\selectfont AIME} & {\fontsize{8}{9}\selectfont HMMT} & {\fontsize{8}{9}\selectfont AIME} & \multirow{2}{*}{Avg.} & \multicolumn{4}{c}{\fontsize{8}{9}\selectfont\textit{LiveCodeBench v6 (code)}} \\
 & 24 & 25 & 25 & 26 &  & 24 & 25 & 25 & 26 &  & {\fontsize{8}{9}\selectfont pass@1} & {\fontsize{8}{9}\selectfont pass@5} & {\fontsize{8}{9}\selectfont pass@8} & {\fontsize{8}{9}\selectfont pass@12} \\
\midrule
Base & 75.8 & 68.1 & 44.7 & 64.4 & 63.3 & 72.2 & 68.1 & 43.3 & 65.6 & 62.3 & 49.6 & 61.7 & 64.1 & 65.7 \\
SFT & 73.6 & 61.1 & 38.9 & 65.8 & 59.9 & 71.7 & 66.9 & 41.1 & 64.4 & 61.0 & 48.5 & 60.7 & 63.3 & 65.1 \\
GRPO & 77.2 & 70.6 & 47.8 & 70.0 & 66.4 & \best{76.1} & 68.9 & 46.4 & 67.2 & 64.7 & 48.7 & 61.0 & 64.4 & 67.4 \\
OPSD & 77.5 & \best{71.7} & 43.9 & 70.8 & 66.0 & 75.6 & \best{69.7} & 46.1 & 69.4 & 65.2 & 49.6 & 60.8 & 63.8 & 66.3 \\
RLCSD & 76.7 & 69.2 & 45.8 & 70.0 & 65.4 & 75.8 & 66.4 & 46.1 & 65.8 & 63.5 & 49.7 & 63.0 & 66.1 & 68.6 \\
\rowcolor{black!6}
ACSD & \best{79.7} & \best{71.7} & \best{49.2} & \best{71.4} & \best{68.0} & 75.6 & 68.9 & \best{47.2} & \best{69.7} & \best{65.3} & \best{51.0} & \best{65.0} & \best{68.4} & \best{70.9} \\
\bottomrule
\end{tabular}
\end{table}

\noindent\textbf{Models and Data.}
Mathematical reasoning experiments cover DeepSeek-R1-0528-Qwen3-8B,
DeepSeek-R1-Distill-Qwen-7B, OLMo-3-7B-Think, Qwen3-8B, and Qwen3-4B
\citep{guo2025deepseekr1,ettinger2025olmo3,qwenteam2025qwen3}; code
generation, ablations, and further analyses use DeepSeek-R1-0528-Qwen3-8B.
The main mathematical comparisons use the OpenThoughts-Math-30K split
released with OPSD \citep{zhao2026opsd}.
For these comparisons, ACSD extracts each direction from 2,048 trajectories on 512 problems and
uses a separate calibration pool for recovery evaluation. Code experiments
train on Codeforces, with directions extracted from a separate pool of
verifier-labeled Codeforces trajectories. Direction-transfer experiments
use DeepMath subsets of different difficulty levels \citep{he2025deepmath}.

\noindent\textbf{Baselines and Training.}
Both tasks compare ACSD with the base model, SFT on reference trajectories,
GRPO with binary outcome rewards, OPSD with a reference-conditioned
self-teacher, and RLCSD combining contrastive self-distillation with
reinforcement learning \citep{shao2024deepseekmath,zhao2026opsd,pan2026rlcsd}.
For mathematics, SFT, GRPO, OPSD, and ACSD use the same framework and LoRA
for 200 updates; RLCSD uses its released implementation for 450 updates.
ACSD and OPSD use a 4,096-token training window; GRPO and RLCSD use 16,384 tokens.
ACSD and OPSD share the forward KL objective with pointwise clipping over
vocabulary entries. Full settings appear in \Cref{app:protocols}.

\noindent\textbf{Evaluation.}
We evaluate mathematical reasoning on AIME24, AIME25, HMMT25, and AIME26. For
each problem, we sample 12 responses and report their average correctness
(Avg@12), together with the average over the four benchmarks. For code
generation, we sample 12 responses per problem on the 175 problems in
LiveCodeBench v6 and report pass@1, pass@5, pass@8, and pass@12
\citep{jain2025livecodebench}. Within each model, all methods share the evaluation generation
budget, sampling parameters, answer parser, and verifier. ACSD evaluates the
distilled student.

\subsection{Main Results}
\label{sec:results}

\noindent\textbf{Mathematical Reasoning.}
The strongest baseline varies across the five models in \Cref{tab:main},
while ACSD achieves the highest average accuracy on each. On
DeepSeek-R1-0528-Qwen3-8B, average accuracy increases from 67.6 to 71.9,
compared with 70.8 for GRPO and 69.0 for both OPSD and RLCSD. On Qwen3-8B,
ACSD reaches 68.0, compared with 63.3 for the base model and 66.4 for GRPO.
ACSD also consistently outperforms OPSD, which shares its distillation
objective and training window.

\noindent\textbf{Extension to Code Generation.}
To evaluate ACSD beyond mathematical reasoning, we construct a self-teacher
from Codeforces trajectories and test code generation on LiveCodeBench v6.
ACSD achieves the highest scores across all four metrics in the code block
of \Cref{tab:main}, increasing pass@1 from 49.6 to 51.0 and pass@12
from 65.7 to 70.9. GRPO and OPSD improve pass@12 while leaving pass@1 at
or below the base model, whereas ACSD improves both single-sample accuracy
and the probability of solving a problem with multiple samples.

\noindent\textbf{Training Cost.}
On Codeforces, ACSD and OPSD use 59.93 and 61.82 H100 GPU-hours for 200
updates; GRPO and RLCSD use 101.21 and 642.46 GPU-hours for 450 updates under
their own protocols, so ACSD obtains its stronger code results at lower
training cost. ACSD additionally spends 10.85 GPU-hours generating the
discovery pool, 2.09 extracting directions at five layers, and 6.17
generating the calibration pool, 79.04 GPU-hours in total with
training, before layer and strength selection; a compact search over a
middle layer and three or four strengths adds to this cost.

\subsection{Ablation Studies}
\label{sec:direction-controls}

\begin{figure}[t]
\centering
\includegraphics[width=0.8\linewidth]{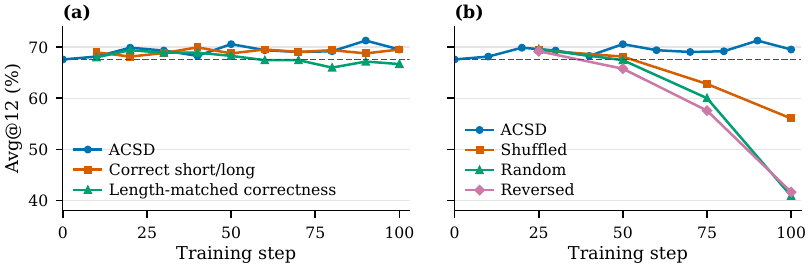}
\caption{Direction ablations on DeepSeek-R1-0528-Qwen3-8B: mean Avg@12 over
four mathematical benchmarks during the first 100 updates. (a) Trajectory
contrasts. (b) Random, shuffled, and reversed directions. Controls share the
layer and strength of the ACSD run in \Cref{tab:main}; dashed lines mark the
base model.}
\label{fig:direction-controls}
\end{figure}

\noindent\textbf{Trajectory Contrasts.}
ACSD groups trajectories by correctness and generation budget. To
examine these choices, we first contrast correct trajectories completed within
and beyond the budget. This direction improves the student throughout
training (\Cref{fig:direction-controls}(a)). These gains show that differences
among successful trajectories with identical correctness labels can provide
useful supervision, supporting the motivation in \Cref{sec:reasoning-behaviors}.
Full ACSD reaches a higher peak accuracy, suggesting that retaining incorrect
and unfinished trajectories adds complementary information. We also contrast
correct and incorrect trajectories matched by length. This direction improves
accuracy initially, then falls below the base model. Together, these results
support contrasting successful with unsuccessful trajectories, with the
budget further separating trajectories that reach the same correct outcome.

\noindent\textbf{Direction Perturbations.}
We replace the ACSD direction with a random unit vector, a
coordinate-shuffled vector, or its reverse. Shuffling preserves the
coordinate values and norm; reversing preserves the axis and changes its
sign. All three perturbations lead to declining accuracy as training
continues, while the original direction consistently improves the student
(\Cref{fig:direction-controls}(b)). The gains thus depend on the extracted
direction and its orientation (construction details in
\Cref{app:direction-construction}).

\subsection{Token-Level Supervision from the Teacher}
\label{sec:token-audit}
\label{sec:length-signal}

The direction ablations establish that activation differences between
trajectories can support self-distillation. To examine how these differences
affect teacher supervision, we fix the student and 512 of its reasoning
trajectories, split evenly into discovery and held-out sets. Both teachers
compute next-token distributions on the same problems and student prefixes:
ACSD applies the direction to hidden states, and OPSD adds a reference
solution to the teacher context. We call the negative gradient of the clipped
distillation loss with respect to student logits the \emph{logit update}.
Positive and negative coordinates indicate upward and downward updates in
logit space. All results below use the 256 held-out trajectories.

\begin{figure}[t]
\centering
\includegraphics[width=0.8\linewidth]{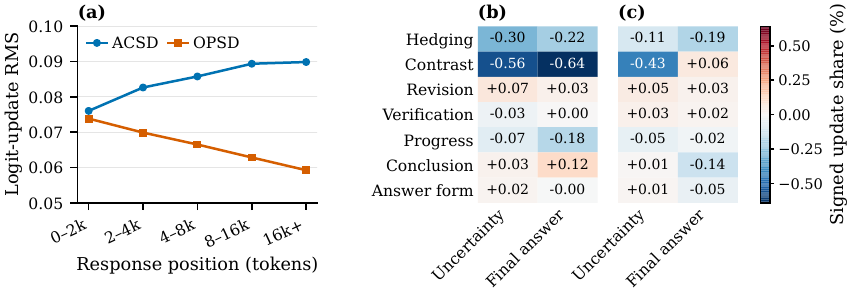}
\caption{Token-level updates on 256 held-out DeepSeek-R1-0528-Qwen3-8B
trajectories under the training clipping rule. (a) Root-mean-square
logit-update magnitude by response position. (b) ACSD and (c) OPSD signed
updates for uncertainty estimation and final answers, as a percentage of the
role's total absolute update; both teachers share prefixes, marker groups, and
color scale.}
\label{fig:teacher-supervision}
\end{figure}

\noindent\textbf{Update Magnitude over Long Trajectories.}
We group prefixes by absolute token position and compute the root-mean-square
norm of the logit-update vectors in each group. In
\Cref{fig:teacher-supervision}(a), ACSD's update magnitude increases early
and then remains relatively stable, while OPSD's decreases as the prefix
grows. OPSD has approximately 97\% of ACSD's magnitude in the first 2k tokens
and 66\% beyond 16k tokens, so ACSD maintains more stable update magnitudes
toward the end of long trajectories. A training-window experiment also finds that ACSD improves
over the base model at every tested window from 1k to 16k tokens; full curves
appear in \Cref{app:training-windows}.

\noindent\textbf{Updates across Reasoning Roles.}
We next examine updates in different reasoning contexts. A language-model
annotator assigns spans to seven roles, including deduction, uncertainty
estimation, backtracking, and final answers, using the response text and role
definitions. We define seven marker groups on the discovery set. Within each
role, we sum the signed updates for each marker group and normalize by the
total absolute update over the vocabulary. \Cref{fig:teacher-supervision}(b--c)
shows uncertainty estimation and final answers, using the same marker groups
for both teachers.

During uncertainty estimation, ACSD assigns negative updates to hedging
markers such as \emph{probably} and \emph{perhaps} and contrast markers such
as \emph{But} and \emph{However}, while assigning positive updates to
revision markers such as \emph{wait} and \emph{Actually}. In final-answer
spans, ACSD assigns negative updates to contrast markers and positive
updates to conclusion markers such as \emph{Therefore} and \emph{Thus}; OPSD
shows the opposite signs for these two groups. ACSD's update size also depends
on the role: the conclusion-marker update is larger in final-answer spans than
during uncertainty estimation, so a fixed vector yields role-dependent
supervision. Full role heatmaps, annotation details, and robustness
checks using an expanded marker lexicon, leave-one-marker-out variants, and
matched random tokens appear in \Cref{app:token-audit}.

\subsection{The Effect of Direction Extraction Data}
\label{sec:transfer}

\noindent\textbf{Direction Transfer.}
We vary the extraction problems to examine the effect of the discovery
distribution (\Cref{tab:extraction-data}(a)). We first retain the OpenThoughts-Math-30K direction and
distill on hard DeepMath problems (Level 7--10), reaching 70.1 Avg@12
compared with the base model's 67.6. We then extract directions from easy
DeepMath problems (difficulty at most 3) and hard problems, and distill on
the same hard set with the model, layer, strength, and optimizer fixed.
These directions reach 70.6 and 71.0, respectively, showing that trajectory
contrasts from easy problems can support self-distillation on harder problems.

\noindent\textbf{Discovery-Set Size.}
We further reduce the discovery set from 512 to 16 problems using nested
subsets, with four trajectories per problem. \Cref{tab:extraction-data}(b)
reports the extraction budget, direction similarity, and distillation result
for each subset.
Every setting improves over the base model. A direction extracted from just
64 trajectories on 16 problems
reaches 70.8 after distillation, compared with 67.6 for the base model, so a
small set of the model's own trajectories suffices to construct an effective
self-teacher (details in \Cref{app:discovery-data}).

\begin{table}[htbp]
\centering
\caption{Direction extraction data on DeepSeek-R1-0528-Qwen3-8B (Avg@12, \%,
over four mathematical benchmarks). (a) Direction transfer across extraction
and training data. (b) Discovery-set size, with four trajectories per problem.
Cosine is measured against the layer-18 direction of the shaded main setting
(512 OpenThoughts-Math-30K problems, $B=8{,}192$).}
\label{tab:extraction-data}
\small
\setlength{\tabcolsep}{4pt}
\begin{minipage}[t]{0.585\linewidth}
\centering
\textbf{(a) Direction transfer}\\[3pt]
\renewcommand{\arraystretch}{1.22}
\begin{tabular}[t]{@{}llrr@{}}
\toprule
Direction data & Training data & Cosine & Avg@12 \\
\midrule
\multicolumn{2}{@{}l}{Base model} & --- & 67.6 \\
\rowcolor{black!6}
OpenThoughts & OpenThoughts & 1.000 & \best{71.9} \\
OpenThoughts & DeepMath (7--10) & 1.000 & 70.1 \\
DeepMath ($\leq 3$) & DeepMath (7--10) & 0.949 & 70.6 \\
DeepMath (7--10) & DeepMath (7--10) & 0.711 & 71.0 \\
\bottomrule
\end{tabular}
\end{minipage}\hfill
\begin{minipage}[t]{0.385\linewidth}
\centering
\textbf{(b) Discovery-set size}\\[3pt]
\renewcommand{\arraystretch}{1.02}
\begin{tabular}[t]{@{}rrrr@{}}
\toprule
Problems & $B$ & Cosine & Avg@12 \\
\midrule
\rowcolor{black!6}
512 & 8,192 & 1.000 & 71.9 \\
256 & 9,216 & 0.951 & 70.4 \\
128 & 8,192 & 0.954 & \best{72.1} \\
64 & 9,216 & 0.935 & 71.5 \\
32 & 9,216 & 0.841 & 71.0 \\
16 & 8,192 & 0.824 & 70.8 \\
\bottomrule
\end{tabular}
\end{minipage}
\end{table}

\FloatBarrier

\section{Related Work}
\label{sec:related}

\noindent\textbf{On-Policy Distillation and Self-Distillation.}
MiniLLM uses reverse KL \citep{gu2024minillm}, and DistiLLM combines skew KL
with adaptive reuse of student-generated responses \citep{ko2024distillm}.
OPSD conditions the teacher on a
reference solution \citep{zhao2026opsd}, while SD-Zero trains a reviser using
correctness feedback and distills its predictions \citep{he2026sdzero}.
RLCSD contrasts correct and incorrect sibling rollouts to modulate the GRPO
advantage \citep{pan2026rlcsd}; OGLS-SD derives teacher-logit adjustments from
successful and failed rollout contexts \citep{oglssd2026}. Other variants
control reference exposure, learn soft prompts, or remove reference-induced
log-probability terms
\citep{han2026atesd,ma2026promptsd,shen2026purifiedopsd}.

\noindent\textbf{Activation Steering for Reasoning.}
Representation engineering studies how activation directions encode and
control model behavior \citep{zou2023repe}; inference-time intervention
improves truthfulness by shifting selected attention-head activations
\citep{li2023iti}. Hidden-state probes also predict the correctness of
intermediate answers \citep{zhang2025reasoningknows}.
Reasoning-specific directions capture execution and reflection or behaviors
such as uncertainty estimation, backtracking, and example testing
\citep{chen2025seal,venhoff2025reasoning}; SEAL and Efficiency Steering apply
them during decoding \citep{chen2025seal,zhao2025efficiency}, and S3-CoT uses
a direction to collect variable-length trajectories for supervised
fine-tuning \citep{du2026s3cot}.
SafeSteer constructs a refusal teacher from activation differences between
harmful and harmless instructions, then restricts reverse-KL distillation to a
global set of selected safety tokens \citep{li2026safesteer}. ACSD constructs
its direction from verifier-labeled reasoning rollouts and provides token-level
supervision throughout complete student-generated rollouts on reasoning tasks.

\section{Discussion}
\label{sec:discussion}

ACSD shows how trajectory-level verification can support token-level
supervision through activation differences. The correct short--long ablation
shows that useful supervision can come from differences among successful
trajectories that share a correctness label, with the generation budget
selecting this contrast; recovery evaluation shows that the intervention
improves continuation success under a fixed budget, and distillation retains
the gains in the student's parameters. The construction fixes one direction,
layer, and strength per training setting. Adapting the intervention to the
prefix, evaluating code generation on further model families, and defining
contrasts from richer execution feedback are natural extensions.

\bibliography{references}
\bibliographystyle{references}

\appendix
\crefalias{section}{appendix}
\crefalias{subsection}{appendix}
\crefalias{subsubsection}{appendix}
\section{Experimental Details and Supplementary Results}
\label{app:protocols}

\subsection{Experimental Setup}

For the results in \Cref{tab:main}, we follow the
checkpoint selection protocol of OPSD \citep{zhao2026opsd}.

\paragraph{Direction Extraction.}
For each model, we select 512 problems from the OpenThoughts-Math-30K split
released with OPSD and sample four closed-book trajectories per problem,
yielding 2,048 trajectories. The split is available as
\path{siyanzhao/Openthoughts_math_30k_opsd}. We extract the last balanced
\texttt{\textbackslash boxed\{\}} expression from each response and compare it
with the dataset answer using \texttt{math\_verify}. When symbolic parsing
fails, we use normalized exact matching. Trajectories with verified final
answers and at most $B$ generated tokens enter the positive group; the rest
enter the contrast group.

The budget for each model is determined from the analysis in
\Cref{sec:reasoning-behaviors}. We extract activations from the first
$\min(|r_i|,B)$ generated tokens of each trajectory, normalize each activation
to unit norm, and pool them by group. Normalizing the difference between the
two group means gives the direction at that layer. \Cref{tab:lcsd-configs}
gives the settings for the main mathematical reasoning experiments. Layer
indices are zero-based.

\begin{table}[htbp]
\centering
\caption{ACSD settings for the main mathematical reasoning experiments.
$B$ is used for trajectory labeling and activation extraction; $\alpha$ is
the intervention strength relative to the current hidden-state norm.}
\label{tab:lcsd-configs}
\small
\setlength{\tabcolsep}{5pt}
\begin{tabular}{@{}lrrrr@{}}
\toprule
Model & $B$ & Layer & $\alpha$ & Training window \\
\midrule
DS-R1-0528-Qwen3-8B & 8,192 & 18 & 0.30 & 4,096 \\
DS-R1-Distill-Qwen-7B & 4,096 & 14 & 0.30 & 4,096 \\
OLMo-3-7B-Think & 4,096 & 16 & 0.10 & 4,096 \\
Qwen3-8B & 4,096 & 13 & 0.05 & 4,096 \\
Qwen3-4B & 4,096 & 13 & 0.05 & 4,096 \\
\bottomrule
\end{tabular}
\end{table}

\paragraph{Optimization.}
In the main mathematical reasoning experiments, SFT, GRPO, OPSD, and ACSD use
the public OPSD framework. We train for 200 updates on eight H100 GPUs,
saving a LoRA adapter every 25 updates. We use AdamW with learning rate
$5\times10^{-6}$, linear decay, effective batch size 32, and maximum gradient
norm 0.1. LoRA is applied to the query, key, value, output, gate, up, and down
projections with rank 64 and LoRA alpha 128.

ACSD and OPSD use distillation temperature $T_{\mathrm d}=1.1$ and pointwise
clipping threshold $\tau=0.06$. ACSD uses the windows in
\Cref{tab:lcsd-configs}; OPSD uses 4,096 tokens. DeepSeek and Qwen student
rollouts use temperature 1.1, top-$p$ 0.95, and top-$k$ 20. OLMo uses
temperature 0.6, top-$p$ 0.95, and no top-$k$ truncation.

SFT trains on reference reasoning trajectories with a 16,000-token window.
GRPO samples eight responses per problem with a 16,384-token window and uses
the answer verifier to assign binary rewards; its KL coefficient is zero.
RLCSD uses its released implementation for 450 updates on the same training
split. It samples eight responses per problem, uses learning rate
$10^{-6}$ with 50 warmup updates and a 16,384-token window, and saves
checkpoints every 50 updates. Four negative hints are selected from the
eight responses to the same problem.

\paragraph{Teacher Computation and Student Updates.}
For each update, the student first generates a response. The student and
frozen teacher then compute next-token distributions on the same student
prefixes. ACSD applies the direction starting at the final prompt state,
which predicts the first response token, and continues at all subsequent
states that predict response tokens. The teacher parameters, direction,
layer, and strength remain fixed. We treat the sampled token sequence and
teacher distributions as constants and update the student using the clipped
objective in \Cref{sec:latent-teacher}. Subsequent rollouts use the updated
student, which is also used for benchmark evaluation.
\Cref{alg:lcsd} summarizes the procedure.

\begin{algorithm}[t]
\caption{Activation-Conditioned Self-Distillation}
\label{alg:lcsd}
\begin{algorithmic}[1]
\Require Base model $\pi_0$, problems $\mathcal D$, verifier $V$, budget $B$,
candidate layers $\mathcal L$, strengths $\mathcal A$,
distillation temperature $T_{\mathrm d}$, clipping threshold $\tau$
\Statex \textbf{Direction discovery}
\State Generate closed-book rollouts $r_i\sim\pi_0(\cdot\mid x_i)$
\State Label $u_i\gets\mathbf 1\{V(x_i,r_i)=1\ \mathrm{and}\ |r_i|\leq B\}$
\State Collect activations within $B$, normalize each to unit norm,
and pool them by $u_i$
\State Compute $\vd^{(\ell)}$ for each $\ell\in\mathcal L$ using
\Cref{eq:direction}
\State Choose $(\ell^\ast,\alpha)$ by recovery on the calibration pool
\Statex \textbf{On-policy self-distillation}
\State Initialize $\pi_\theta\gets\pi_0$ and freeze a copy of $\pi_0$
as the teacher $\widetilde\pi_0$
\For{each update}
  \State Sample $x\sim\mathcal D$
  \State Sample unsteered $y\sim\pi_\theta(\cdot\mid x)$
  \State Compute $p_t^S$ and $p_t^T$ on $(x,y_{<t})$ using
  \Cref{eq:steering-background,eq:teacher-student}
  \State Compute $\ell_{t,v}$ using \Cref{eq:pointwise-kl}
  \State Update $\theta$ with the clipped objective in \Cref{eq:lcsd}
\EndFor
\State Discard the frozen teacher and $\vd$
\end{algorithmic}
\end{algorithm}

\paragraph{Mathematical Evaluation.}
\label{app:measurement}
We sample 12 responses per problem on AIME 2024, AIME 2025, HMMT February
2025, and AIME 2026, 30 problems each (Hub identifiers
\path{HuggingFaceH4/aime_2024}, \path{MathArena/aime_2025},
\path{MathArena/hmmt_feb_2025}, and \path{MathArena/aime_2026}).
\Cref{tab:eval-sampling} lists the
decoding settings for each model family. Methods using the same model share
these settings, the answer parser, and the verifier. Min-$p$ and the presence
penalty are zero. DeepSeek-family generation uses a 40,960-token context and
begins reasoning with \texttt{<think>}; Qwen3 uses thinking mode.

\begin{table}[htbp]
\centering
\caption{Decoding settings for mathematical reasoning benchmarks.}
\label{tab:eval-sampling}
\small
\setlength{\tabcolsep}{5pt}
\begin{tabular}{@{}lrrrr@{}}
\toprule
Model family & Max tokens & Temperature & Top-$p$ & Top-$k$ \\
\midrule
DeepSeek-R1 & 38,912 & 0.6 & 0.95 & -- \\
OLMo-3-Think & 38,912 & 0.6 & 0.95 & -- \\
Qwen3 & 38,912 & 1.0 & 0.95 & 20 \\
\bottomrule
\end{tabular}
\end{table}

\paragraph{Code Generation.}
Code experiments use 5,000 Codeforces problems from the Open-R1 release
(\path{open-r1/codeforces}, configuration \texttt{verifiable-prompts}),
restricted to problems with a Python prompt. ACSD trains on the problem
statements alone; SFT and OPSD use the Python reference implementations of
the same problems from \path{open-r1/codeforces-cots}. The direction is
extracted from a separate pool of trajectories: four closed-book trajectories
for each of 512 Codeforces problems, generated with a 32,768-token limit and
labeled by the official tests, with $B=16{,}384$, layer 18, and
$\alpha=0.10$. ACSD and OPSD use the optimizer, LoRA settings, sampler, and
clipping of the mathematical experiments with a 4,096-token window and 200
updates, saving an adapter every 25 updates; GRPO and RLCSD train for 450
updates with a 16,384-token window, RLCSD with full-parameter updates.
Evaluation uses the 175 problems of LiveCodeBench v6
(\path{livecodebench/code_generation_lite}) with 12 samples per problem,
temperature 0.6, top-$p$ 0.95, no top-$k$ truncation, and up to 32,768
generated tokens, scored with the official checker.

\FloatBarrier
\subsection{Recovery Evaluation and Configuration Selection}
\label{app:recovery}
\label{sec:selection}

Recovery evaluation uses the calibration pool: 256 problems held out from
direction extraction.
We shuffle them with seed 17, select one eligible contrast-group trajectory
per problem, and assign prefix fractions of 25\%, 50\%, and 75\% in
round-robin order. The initial DeepSeek-R1-0528-Qwen3-8B scan uses the full
contrast group. Later model scans use contrast-group trajectories that
terminate naturally and contain a boxed answer.

We continue each fixed prefix with and without the direction, using the
same sampling settings and continuation budget across configurations. Each
prefix receives two continuations, each capped at 12,288 generated tokens.
Recovery is the fraction of continuations whose final answers pass
verification, with 512 continuations per layer--strength setting. DeepSeek
and Qwen use temperature 1.1, top-$p$ 0.95, and top-$k$ 20. OLMo uses
temperature 0.6, top-$p$ 0.95, and no top-$k$ truncation.

\begin{figure}[htbp]
\centering
\includegraphics[width=0.94\linewidth]{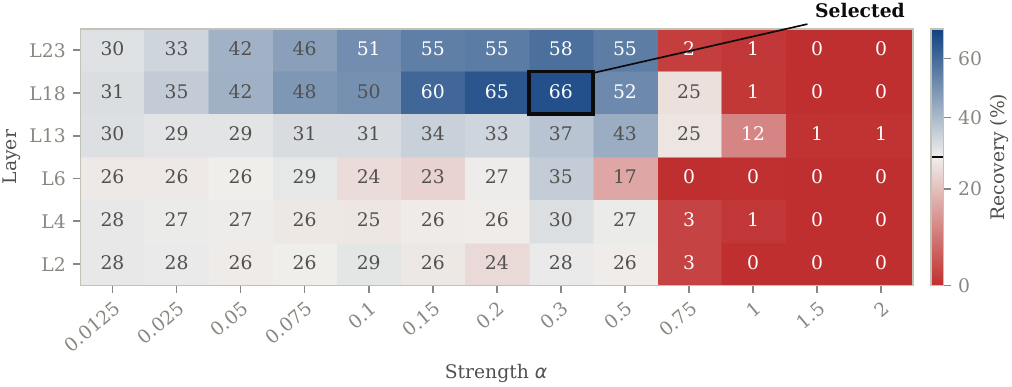}
\caption{Recovery scan on DeepSeek-R1-0528-Qwen3-8B. Each cell reports the
percentage of 512 continuations that pass answer verification. The black
outline marks layer 18 and $\alpha=0.3$, used for distillation. Recovery
without steering is 26.6\%.}
\label{fig:probe}
\end{figure}

The initial scan on DeepSeek-R1-0528-Qwen3-8B includes layers
$\{2,4,6,13,18,23\}$ and strengths
$\{0.0125,0.025,0.05,0.075,0.1,0.15,0.2,0.3,0.5,0.75,1,1.5,2\}$.
\Cref{fig:probe} shows higher recovery in intermediate layers. The selected
layer 18 and $\alpha=0.3$ raise recovery from 26.6\% to 66.0\%. Later models
use five candidate layers and $\alpha\in\{0.05,0.1,0.2,0.3\}$. The layer
grids are $\{6,10,14,18,22\}$ for DeepSeek-R1-Distill-Qwen-7B,
$\{6,11,16,20,25\}$ for OLMo-3-7B-Think, and $\{6,13,18,23,29\}$ for both
Qwen3 models. Trajectory labels and directions at each layer are fixed
before the scan; recovery selects the intervention layer and strength.

\paragraph{Downstream Sensitivity.}
\label{app:selection-sensitivity}
We compare six intervention configurations in distillation: at layer 18,
$\alpha\in\{0.05,0.3,0.75\}$; at $\alpha=0.3$, layers 4, 13, 18, and 23.
All settings share the model, training data, optimizer, and evaluation
protocol. The recovery-selected setting reaches 71.9 Avg@12, while the five
alternatives range from 68.7 to 69.9.

\begin{figure}[htbp]
\centering
\includegraphics[width=0.72\linewidth]{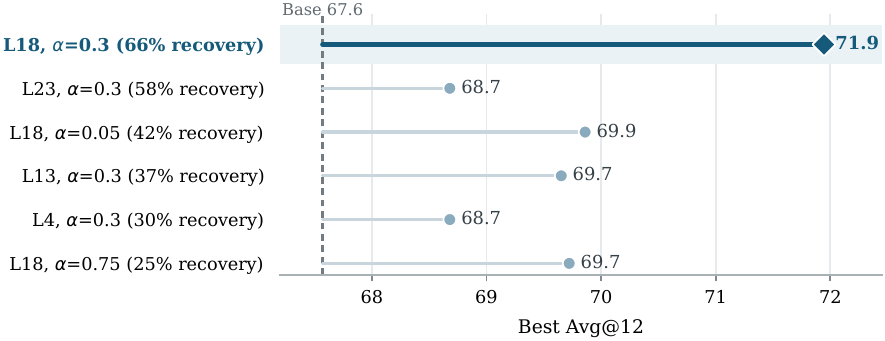}
\caption{Distillation results for six intervention settings on
DeepSeek-R1-0528-Qwen3-8B. The horizontal axis shows the best mean Avg@12
across four mathematical reasoning benchmarks. Labels also give recovery;
the outline marks the main setting.}
\label{fig:selection-sensitivity}
\end{figure}

\FloatBarrier
\subsection{Complete Direction Controls}
\label{app:direction-construction}
\label{app:additional-controls}

\paragraph{Trajectory Contrasts.}
The correct short--long direction contrasts 914 correct trajectories that
finish within the generation budget with 913 correct trajectories that
finish beyond it. The length-matched correctness direction contrasts all
221 incorrect trajectories with 221 correct trajectories selected by
nearest-neighbor matching on response length. Their mean lengths differ by
78 tokens. All groups use the same activation normalization and
difference-in-means construction.

\paragraph{Direction Perturbations.}
The random direction is an isotropic unit vector. The shuffled direction
randomly permutes the coordinates of the ACSD direction, and the reversed
direction is $-\vd$. These controls share ACSD's intervention layer,
strength, training data, and optimizer.
\Cref{fig:direction-controls} shows the trajectory and perturbation comparisons.

\FloatBarrier
\subsection{Token-Level Update Analysis}
\label{app:token-audit}

\paragraph{Fixed-Trajectory Evaluation.}
We use 512 fixed student trajectories from DeepSeek-R1-0528-Qwen3-8B, one
per problem, split into 256 discovery and 256 held-out trajectories. The
base model, ACSD teacher, and OPSD teacher process the same problem and
student prefix at every position. Marker groups are determined on the
discovery set, and measurements are reported on the held-out set. The role
analysis uses full held-out trajectories to retain less frequent roles.

\paragraph{Update Magnitude.}
At prefix $s$, let $\vu_s$ denote the negative gradient of the clipped
distillation loss with respect to student logits, with $u_s(v)$ denoting its
coordinate for vocabulary entry $v$. We group prefixes by
absolute response position into five bins: 0--2k, 2--4k, 4--8k, 8--16k,
and beyond 16k tokens. For bin $b$ containing $N_b$ prefixes, the update
magnitude is
\begin{equation}
R_b=\sqrt{\frac{1}{N_b}\sum_{s\in b}\lVert\vu_s\rVert_2^2}.
\label{eq:update-rms}
\end{equation}
Both teachers compute updates on the same observed text. OPSD's magnitude
relative to ACSD is 97\% in the first 2k tokens, 85\% in the next 2k, and
66\% beyond 16k. Over the same range, the cosine similarity between their
update vectors falls from 0.298 to 0.108.

\paragraph{Role Annotation.}
We adapt the reasoning-behavior taxonomy of \citet{venhoff2025reasoning},
adding a \texttt{final-answer} category; spans are segmented and annotated
with the protocol described below. We split trajectories at sentence and
paragraph boundaries, with at most 900 characters per span. The annotator, \texttt{deepseek-chat}, uses
temperature 0 and assigns one role per span from the response text and the
definitions in \Cref{tab:role-definitions}. Malformed or missing outputs are
marked unlabeled and excluded from role aggregates.

\begin{table}[htbp]
\centering
\caption{Reasoning roles used in the token-level audit.}
\label{tab:role-definitions}
\small
\setlength{\tabcolsep}{4pt}
\begin{tabular}{@{}p{0.18\linewidth}p{0.24\linewidth}p{0.48\linewidth}@{}}
\toprule
Figure label & Annotator label & Definition \\
\midrule
Setup & \texttt{initializing} & Restating the task, setting up variables, or planning. \\
Deduction & \texttt{deduction} & Deriving a conclusion from the current assumptions. \\
Known facts & \texttt{adding-knowledge} & Recalling a theorem, formula, identity, or external fact. \\
Examples & \texttt{example-testing} & Checking a case, example, or sanity check. \\
Uncertainty & \texttt{uncertainty-}\newline\texttt{estimation} & Expressing uncertainty, hedging, or assessing confidence. \\
Backtracking & \texttt{backtracking} & Abandoning, revising, or correcting an approach. \\
Final answer & \texttt{final-answer} & Presenting, boxing, or restating the final answer. \\
\bottomrule
\end{tabular}
\end{table}

The annotation instruction is reproduced below.

{\footnotesize
\begin{verbatim}
You are annotating reasoning traces for a mechanistic analysis.
Label each span with exactly one label. Do not merge or drop spans.

Allowed labels:
- initializing: restating the task, setting up variables, or planning
- deduction: deriving a conclusion from current assumptions
- adding-knowledge: recalling a theorem, formula, identity, or fact
- example-testing: checking a case, example, or sanity check
- uncertainty-estimation: expressing uncertainty, hedging, or confidence
- backtracking: abandoning, revising, or correcting an approach
- final-answer: presenting, boxing, or restating the final answer

Return strict JSON mapping every span id to one allowed label.
\end{verbatim}
}

\paragraph{Marker Groups and Aggregation.}
We use the seven groups of single-token markers in
\Cref{tab:marker-groups}. Leading-space variants are retained when the
tokenizer maps them to distinct single tokens. Marker groups and reasoning
role labels are determined separately.

\begin{table}[htbp]
\centering
\caption{Marker groups used in the token-level update analysis.}
\label{tab:marker-groups}
\small
\setlength{\tabcolsep}{6pt}
\begin{tabular}{@{}ll@{}}
\toprule
Group & Examples \\
\midrule
Hedging & probably, perhaps, maybe, possibly \\
Contrast & But, but, However, however \\
Revision & wait, Wait, Actually, actually, reconsider \\
Verification & check, verify \\
Progress & so, then, Then, implies \\
Conclusion & Therefore, Thus, hence, conclude \\
Answer form & final, Final, answer, Answer, boxed \\
\bottomrule
\end{tabular}
\end{table}

\begin{figure}[htbp]
\centering
\includegraphics[width=\linewidth]{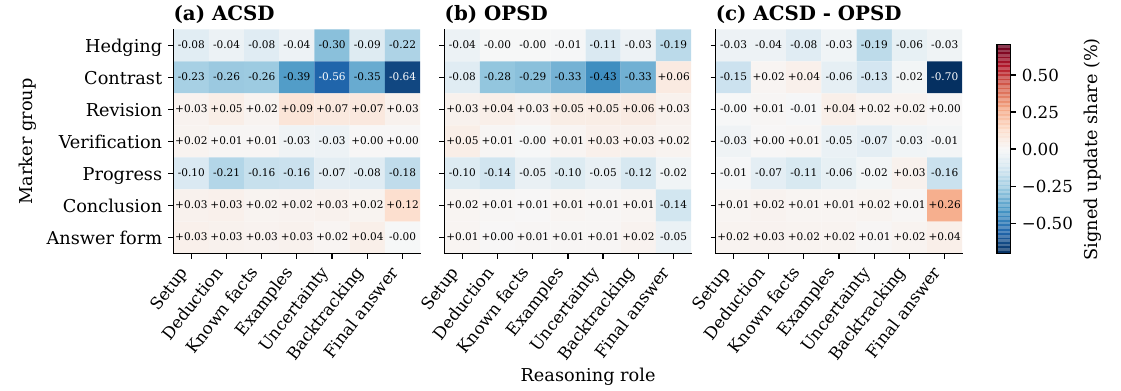}
\caption{Complete role and marker updates on 256 held-out trajectories.
(a) ACSD. (b) OPSD. (c) ACSD minus OPSD normalized updates. Rows
are marker groups; columns are reasoning roles.}
\label{fig:token-audit-full}
\end{figure}

Let $\mathcal S_r$ be the set of held-out prefixes with role $r$ and
$\mathcal V_g$ the vocabulary entries in marker group $g$. The percentage
shown in each heatmap cell is
\begin{equation}
A_{g,r}=100\times
\frac{\sum_{s\in\mathcal S_r}\sum_{v\in\mathcal V_g}u_s(v)}
{\sum_{s\in\mathcal S_r}\lVert\vu_s\rVert_1}.
\label{eq:marker-update}
\end{equation}
\Cref{fig:teacher-supervision} shows uncertainty estimation and final
answers; \Cref{fig:token-audit-full} shows all seven roles. Both teachers
use the same marker groups, role labels, normalization, and color scale.
An expanded marker lexicon and leave-one-marker-out analysis preserve the
main sign patterns described in \Cref{sec:token-audit}. We also compare
random token groups matched by discovery-set occurrence and update scale.
The aggregate EOS update is near zero and slightly negative.

\FloatBarrier
\subsection{Training-Window Experiments}
\label{app:training-windows}

We run a separate 100-update experiment on DeepSeek-R1-0528-Qwen3-8B. ACSD
uses windows of 1k, 2k, 4k, 8k, and 16k tokens; OPSD uses 1k, 2k, 4k, and
8k. Within each method, we hold the model, optimizer, training data, and
evaluation protocol fixed. ACSD also keeps the direction and intervention
settings fixed. Only the training window changes.

\begin{figure}[htbp]
\centering
\includegraphics[width=\linewidth]{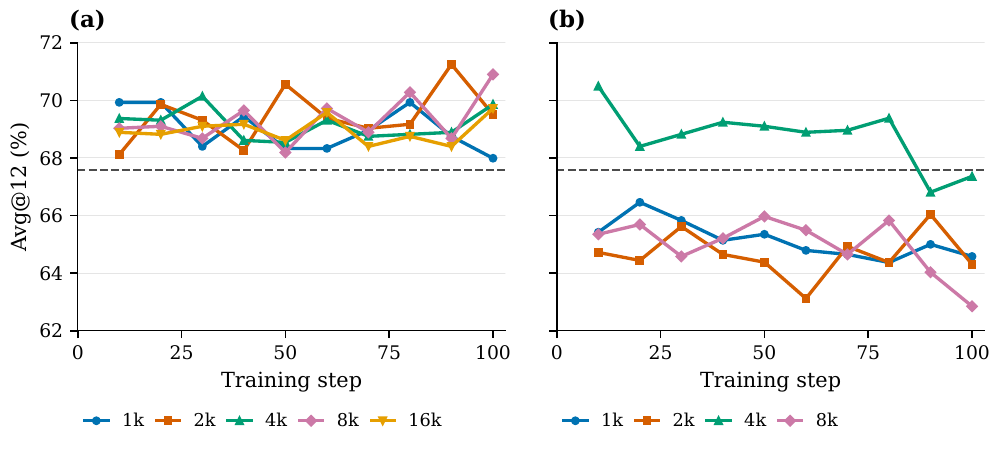}
\caption{Training curves for different windows on
DeepSeek-R1-0528-Qwen3-8B. We report mean Avg@12 across four mathematical
reasoning benchmarks over the first 100 updates. (a) ACSD. (b) OPSD. Lines
connect evaluations; dashed lines mark the base model.}
\label{fig:training-windows}
\end{figure}

\Cref{fig:training-windows} shows the full training curves. ACSD improves
over the base model at every tested window. OPSD at 4k exceeds the base
model early in training and reaches 67.4 at update 100, slightly below the
base model's 67.6. The curves for its other tested windows remain below
the base model.

\subsection{Direction Transfer and Discovery-Set Size}
\label{app:discovery-data}

\paragraph{Transfer across Data and Difficulty.}
The OpenThoughts transfer setting retains the direction extracted from
OpenThoughts-Math-30K and replaces the distillation problems with DeepMath
Level 7--10, using the split released with RLCSD (\path{Leyiii/RLCSD}). In the easy-to-hard comparison, we collect 2,048 trajectories
each from DeepMath problems with difficulty at most 3 and from Level 7--10
problems. Their budgets are 6,144 and 19,456 tokens, yielding 50.6\% and
50.7\% within-budget verified trajectories. Both directions are used for
distillation on the same hard problems, with the same model, intervention
layer, strength, optimizer, and evaluation settings.

\Cref{tab:extraction-data}(a) summarizes the transfer results. At layer 18,
the directions from easy and hard DeepMath problems have cosine similarities
of 0.949 and 0.711 with the OpenThoughts-Math-30K direction, respectively.
These comparisons use the direction employed for distillation in each
setting and its corresponding generation budget.

\paragraph{Discovery-Set Size.}
We use nested subsets of 512, 256, 128, 64, 32, and 16 discovery problems
for DeepSeek-R1-0528-Qwen3-8B, sampling four trajectories per problem.
\Cref{tab:extraction-data}(b) gives the budget, cosine similarity to the
512-problem direction, and distillation result for each subset.
Cosines are computed from the normalized direction vectors at layer 18.
As the discovery set shrinks, we keep layer 18 and $\alpha=0.3$
fixed and use the main mathematical evaluation settings.

\FloatBarrier
\subsection{Inference-Time Steering}
\label{app:inference-steering}

We apply the direction, layer, and strength selected for teacher construction
throughout full autoregressive generation from the base model. The
intervention is active whenever the model predicts a response token.
Evaluation follows the main mathematical decoding settings.
\Cref{tab:inference-steering} reports the results.

\begin{table}[htbp]
\centering
\caption{Direction intervention throughout generation. Values are average
Avg@12 (\%) across four mathematical reasoning benchmarks;
$\Delta$ denotes Steered minus Base. Intervention settings come from
\Cref{tab:lcsd-configs}.}
\label{tab:inference-steering}
\small
\setlength{\tabcolsep}{5pt}
\begin{tabular}{@{}lcrrr@{}}
\toprule
Model & Layer / $\alpha$ & Base & Steered & $\Delta$ \\
\midrule
DS-R1-0528-Qwen3-8B & $(18,0.30)$ & 67.6 & 25.6 & $-42.0$ \\
DS-R1-Distill-Qwen-7B & $(14,0.30)$ & 40.4 & 13.8 & $-26.6$ \\
OLMo-3-7B-Think & $(16,0.10)$ & 64.1 & 62.2 & $-1.9$ \\
Qwen3-8B & $(13,0.05)$ & 63.3 & 62.1 & $-1.2$ \\
Qwen3-4B & $(13,0.05)$ & 62.3 & 62.4 & $+0.1$ \\
\bottomrule
\end{tabular}
\end{table}

This evaluation changes the autoregressive trajectory from the first
generated token. Recovery evaluation starts from a fixed prefix and compares
success under the same continuation budget. The ACSD teacher computes
distribution targets on prefixes already generated by the student, and
final evaluation uses the distilled student. These settings examine full
generation, fixed-prefix continuation, and performance after student
training, respectively.

\end{document}